\documentclass[11pt,letterpaper]{article}
\usepackage{acl_style}
\usepackage{times}
\usepackage{latexsym}

\usepackage{amsmath, amsthm, amssymb}
\usepackage{array}
\usepackage{graphicx}

\title{Good, Better, Best: Choosing Word Embedding Context}

\author{
\vspace{0.3cm}
\hspace{2.5cm}James Cross$^1$ \hspace{2.3cm} Bing Xiang$^2$  \\
$^1$School of EECS\\
Oregon State University\\
Corvallis, OR\\
{\tt crossj@oregonstate.edu}
\And
\vspace{0.3cm}
\hspace{2cm}Bowen Zhou$^2$\\
$^2$IBM Watson\\
T. J. Watson Research Center\\
Yorktown Heights, NY\\
{\tt \{bingxia,zhou\}@us.ibm.com}
}

\date{}

\begin{document}
\maketitle
\begin{abstract}
We propose two methods of learning vector representations of words and phrases that each combine sentence context with structural features extracted from dependency trees.  Using several variations of neural network classifier, we show that these combined methods lead to improved performance when used as input features for supervised term-matching.
\end{abstract}

\section{Introduction}

Continuous-space embeddings of words and phrases are very appealing for many applications in NLP.   When used as a feature to represent words, they are much more compact than ``one-hot" vectors.  In addition, their geometry can encode a wealth of information about relationships between words, and thus their meanings, which can be exploited by downstream applications.

The critical question is how to best achieve that geometry for a given task.  There are two types of training data which are often used to learn word vectors.  Approaches based on the ``distributional hypothesis" assume that the semantics of a word can be determined through the contexts in which it appears.  They model vectors through some form of dimension compression on word co-occurrence statistics taken from an example text.  Other methods learn vectors from pre-existing information about specific relationships between words, such as those found in a structured knowledge base.

We propose two vector-learning methods which combine sentence-context input with structural features extracted automatically from dependency trees for the same underlying corpus.  We compare them with vectors trained from only one of these types of input, and show that they lead to improved performance on simple classification tasks.

\section{Related Work}

Many different approaches to learning vector representations of words have been taken.  Methods trained from large text examples by relying directly or indirectly on word co-occurrence counts date back to latent semantic analysis \cite{Deerwester:90}.  Recent examples include the window-based approach of Mikolov et al. (2013c) and the GloVe method \cite{Pennington:14}.  There has also been work incorporating prior knowledge with bag-of-words context \cite{Yu:14}.

It is also possible to perform unsupervised learning on structured information about the underlying entities or concepts being represented, such as the relationships encoded in an online knowledge base.  Approaches for utilizing this type of training data include the use of neural networks and/or kernel density estimation \cite{Bordes:11} and encoding relations as translations on hyperplanes \cite{Wang:14}.

The most direct precursor to our work is the well-known Word2Vec algorithm \cite{Mikolov:13a,Mikolov:13b,Mikolov:13c}.  It consists of a simple single-layer neural model where the goal is to maximize the ability of the word representations to predict neighbors of the word in a large raw-text corpus.  It was observed by Levy and Goldberg (2014) that this algorithm could be generalized to use any type of discrete features with which words could be paired as the ``contexts'' to be predicted.

\section{Combined Objective}

%In this work, we explore what gains may be made by combining both linear context-window pairings and arbitrary word features.  In particular, we applied two distinct approaches: the joint-objective approach, where both types of training data were used in parallel, and the sequential approach, where vectors were trained using Word2Vec were used to initialize the word/feature pair algorithm.

In this work, we combine sentence-window context with structural features described in Section~\ref{sec:feats}.  The joint objective consists of using both types of training data in parallel so that the training objective is to maximize the combined log probability:

\vspace{-0.5cm}
\begin{equation} \label{joint-prob}
\begin{split} 
  \frac{1}{T} \sum_{t=1}^{T} \sum_{-c \leq j \leq c, j \neq 0} \log p(w_{t+j} | w_t) \\
  + \hspace{0.25cm}  \frac{\alpha}{|P|} \sum_{(w,f) \in P} \log p(f | w)
\end{split}
\end{equation}

\noindent
where $T$ is the number of words in the training corpus, $c$ is the context window size, $P$ is the set of all pairs of words and structural features $(w,f)$, and $\alpha$ is a weighting parameter as between the two types of training data.  

The conditional probabilities are modeled with vectors as in Word2Vec, and two sets of output vectors, internal to the algorithm, are learned to represent words and structural features as contexts.  This objective is implemented with a joint algorithm which performs updates using the sentence context and the word/feature pairs in parallel.  Both types of updates are based on the negative sampling training method \cite{Mikolov:13c}.

We also implemented a sequential algorithm where vectors trained with just sentence-window context are used as initial representations which are subsequently refined by updates using the structural features.  The relative success of vectors trained in this way suggests that proceeding from a more general representation toward one which is more directly tailored for a given task is a promising avenue.

In the sections that follow, we describe a number of simple classification experiments designed to compare sets of word embeddings trained using these various methods.  For the term-matching tasks we considered, the structural features alone led to better-performing vectors than those trained with just linear word context, while the combined-objective methods provided further improvement.

\begin{table}   %\label{tab:patt-feats}
  \centering
  \scalebox{0.8}{
  \begin{tabular}{|l|p{4cm}|p{3cm}|}
    \hline
    {\bf Feature} & {\bf Dependency Pattern} & {\bf Example}\\
    \hline \hline

    {\tt such\_as} & 
    $mwe(\texttt{as}_i,\texttt{such}) \land \newline
    prep(X,\texttt{as}_i) \land \newline
    pobj(\texttt{as}_i,Y)$ & 
    ... {\bf anarchists} such as Frederico {\bf Urales} ... \\   
    \hline

    {\tt known\_as} & 
    $prep(\texttt{known}_i,\texttt{as}_j) \land \newline
    nsubjpass(\texttt{known}_i,X) \land %\newline
    pobj(\texttt{as}_j,Y)$ & 
    The {\bf incident} became known as the Haymarket {\bf affair}. \\   
    \hline
    
    {\tt name\_for} & 
    $nsubj(X,\texttt{name}_i) \land \newline
    prep(\texttt{name}_i,\texttt{for}_j) \land \newline
    pobj(\texttt{for}_j,Y)$ 
    & The trivial (non-systematic) name for {\bf alkanes} is {\bf paraffins}. \\   
    \hline
  \end{tabular}}
  \caption{Examples of noun-relationship pattern features extracted from dependency structures. In each case, one feature is created for word $X$ which includes the feature type and $Y$, and another is created for $Y$ which includes the feature type and $X$.  Subscripts indicate sentence position.}
  \label{tab:patt-feats}

\end{table}

\section{Structural Features} \label{sec:feats}

As a representative means of training word vectors on functional relationships between words in sentences, rather than just word proximity, we began with first-order dependency arcs of the type described in Levy and Goldberg (2013).  For every dependency arc, each word is annotated with a feature which consists of the arc label, whether the word is the head or tail of the arc, and the adjoined word.\footnote{Also included from that work and in the ``Dependency" test set are features coming from the ``flattening'' of prepositional phrases to directly link the modified head and the object of the preposition (with the preposition itself included in the feature name).}  We also extracted two types of higher-order features from the dependency trees: noun-relationship patterns, and subject-object relationships.

Noun-relationship features capture certain taxonomic relationships between nouns, as reflected in the structure of a sentence, such as when one word names an instance of a category defined by another word, or when they are different names for the same thing.  This methodology is based on the syntactic patterns used to recognize {\em is-a} relationships in \cite{Fan:11}.  Examples of these dependency patterns and the relationships they identify can be seen in Table~\ref{tab:patt-feats}.

In addition, second-order features directly linking subjects and objects of transitive verbs (and including the verb) are also extracted.  Thus for the sentence ``The woman ate the paella'' the word ``woman'' will be annotated with a feature indicating that it is the subject of the verb ``ate'' {\em with the direct object ``paella''}, and ``paella'' will receive a feature marking it as the object of ``ate'' with the subject ``woman''.

The rationale for using both of these types of features is that nouns contain much of the semantic nexus of a sentence, and they have direct dependency relationships with essentially all other types of words.  By refining word vectors based on specific relationships between nouns, the quality of the vector representations as a whole can be improved.

The features used to train vectors for our experiments were extracted from dependency trees produced by the Stanford Neural Network Dependency Parser \cite{Chen:14} using the provided pre-trained model with Stanford dependency labels.  The underlying corpus for all methods was the full text of English Wikipedia as of June 15, 2014.\footnote{As maintained by the KOPI project \cite{Pataki:12}.}

\section{Qualitative Comparison}

\begin{table}
  \centering
  \tiny
  \begin{tabular}{| m{1.5cm} || m{1.5cm} | m{1.5cm} | m{1.5cm} |}
  \hline
     {\bf Target Word}  & Raw Text \newline (Word2Vec) & Structural \newline Features & Joint \\
   \hline \hline
%   turing &
%   \shortstack[l]{multi-tape \\ nondeterministic \\ recursion \\ g\"odel \\ multitape} & 
%   \shortstack[l]{hamming \\ heyting \\ stieltjes \\ raphson \\ hawking} & 
%   \shortstack[l]{presburger \\ hamming \\ raphson \\ feynman \\ fredkin} \\
%   \hline
%
%    florida &
%    \shortstack[l]{tallahassee \\ jacksonville \\ brevard \\ broward \\ miami} &
%    \shortstack[l]{texas \\ carolina \\ arizona \\ louisiana \\ california} &
%    \shortstack[l]{carolina \\ arizona \\ texas \\ virginia \\ miami} \\
%    \hline 
%    
    part-of-speech &
    \shortstack[l]{xml-like \\ meta-language \\ sxml \\ script-based \\ canonicalization} &
    \shortstack[l]{file-type \\ non-template \\ single-word \\ xml-style \\ single-letter} &
    \shortstack[l]{verb-noun \\ html-like \\ multiword \\ script-based \\ now-deprecated} \\
    \hline 
    
    dynamic &
    \shortstack[l]{decouples \\ user-centric \\ gesture-based \\ multi-path \\ process-based} &
    \shortstack[l]{spacial [{\em sic}] \\ timbral \\ user-centric \\ task-focused \\ content-driven} &
    \shortstack[l]{nonhierarchical \\ composition-based \\ non-virtual \\ ever-changing \\ human/machine} \\
    \hline
    
    normalize &
    \shortstack[l]{normalise \\ normalizing \\ normalising \\ rebalance \\ normalization} &
    \shortstack[l]{normalise \\ regularize \\ exponentiate \\ recalculate \\ quantify} &
    \shortstack[l]{normalise \\ regularize \\ quantify \\ define \\ homogenize} \\
    \hline        
    \end{tabular}
        
    \caption{Most similar 5 words (by cosine similarity) to several target words under different vector-training regimes.}
    \label{Qualitative}
\end{table}

To evaluate the relative qualities captured by the various embedding methods, we looked at the closest words to a number of different target words, just as in Levy and Goldberg (2014).  Unsurprisingly, the nearest words were very similar across the different vector sets, as all were trained with the same base corpus.  

In particular, jointly-trained vectors lead to similar words which are subjectively very much like those produced by the vectors trained with structural features only.  There is reason to believe, however, that the joint-objective vectors capture some additional semantic nuance stemming from the intersection of (and interplay between) topical and functional similarity.  

For the word ``part-of-speech" the most similar words under all vector sets skew toward hyphenated or compound modifiers (though the set induced by bag-of-words training includes nominals).  While all of the vector sets seem to reflect the fact that this modifier is frequently used with the word ``tag(s)," the jointly-trained vectors also capture its affinity with linguistics in such similar words as ``verb-noun" and ``multiword."  Likewise with other words such as ``dynamic" and ``normalize" the jointly-trained vectors show properties which are similar to those using structural features alone, but also seem to encode meaning in a somewhat more targeted way (``ever-changing"; ``homogenize").

\section{Experiments}

In order to compare their predictive properties quantitatively, we tested word embeddings produced with each method as inputs to binary term-matching classifiers using two different data sets, described below.  We used shallow neural network classifiers with three alternative architectures, with a uniform hidden-layer size of 200. The results are presented in Table~\ref{tab:Results}.

For the standard multilayer perceptron (``MLP") algorithm, the vector representations of each word in a pair were concatenated to make the input to a single hidden layer with a leaky rectifier activation function \cite{Glorot:11,Maas:13}.  We also used a version where the two vectors were each used as input to the same hidden layer, i.e., the weights were shared, and the hidden-layer output for each, together with their element-wise absolute-value difference, were concatenated as input to the logistic regression classifier layer (``Shared MLP").

The third classifier similarly used shared weights to process each vector input (with a hyperbolic tangent activation function), but classification was made based on the cosine similarity of the hidden-layer output, with a threshold of 0.5.  In this way, the network learns how to project the vectors into a space where their angular separation is a good measure of their similarity for the task at hand.  We observed that this network converged very fast during training, but had performance nearly on par with more complicated models.

\begin{table}
  \centering
  \small
  \begin{tabular}{ l | p{1.0cm} | p{1.0cm} | p{1.0cm} }
    \hline
    & MLP & Shared MLP & Shared Cosine \\
    \hline \hline
    \multicolumn{4}{l}{\textbf{WordNet}} \\
    \hline \hline
    Raw Text (Word2Vec) & 75.88 & 73.76 & 77.10 \\
    \hline
    Dependency & 82.71 & 79.56 & 80.74 \\ 
    \hline
    + Stuctural Features & 82.86 & 80.70 & 81.61 \\
    \hline
    Joint & 82.86 & 80.81 & 81.65 \\
    \hline
    Sequential & 83.12  & 81.61  & 82.75 \\
    \hline 

    \hline \hline
    \multicolumn{4}{l}{\textbf{PPDB}} \\    
     \hline \hline
    Raw Text (Word2Vec) &  93.43 & 92.84 & 93.20 \\
    \hline
    Dependency & 94.88 & 94.05 & 94.06 \\
    \hline
    + Structural Features &  94.82 & 94.00 & 94.09 \\
    \hline
    Joint &  95.52 & 94.63 & 94.51 \\
    \hline
    Sequential &  95.48 & 95.03 & 94.53 \\
    \hline    
   
    \end{tabular}
    
        \caption{Accuracy results (\%) for classification of WordNet synonym pairs and PPDB lexical matches for different vector sets}
    \label{tab:Results}
\end{table}

\subsection{WordNet Synonyms}

The first source of term-matching pairs we considered were synonyms from the WordNet lexical database \cite{Miller:95}.  WordNet groups words into synsets which are groups of words (and multi-word phrases) with the same meaning.  We created match data by taking all distinct pairs of words within synsets of any part of speech, and generated negative examples by random shuffling.\footnote{The WordNet data, as filtered for coverage by these vector sets, consists of 21,089 training pairs, 2,636 development pairs, and 2,637 test pairs, split evenly between matches and non-matches.}

This turns out to be a relatively difficult task with unsupervised vectors alone.  A possible reason is that the vectors do not distinguish between word senses, essentially folding together all usage cases into a single representation.  The WordNet data is characterized by relatively common words where the synonymy may rely on a secondary or figurative definition.  Examples of this in our data are the pairs ({\tt fastball}, {\tt bullet}) and ({\tt arrest}, {\tt collar}).  In particular, since the raw-text data tends to group words by domain of discourse rather than function, it makes sense that a figurative usage as for ``bullet" and ``collar" in those pairs might cause problems.

In this case, the structural features were much more useful than bag-of-words context.  The pattern-based features also led to a clear improvement over using just the subset of first-order dependency relations, which we hypothesize is largely due to the subject-object relations.  The joint vectors showed little gain over using structural features only.  This seems linked to the large gap in utility between the two types of training data, and points to the importance of application domain in determining embedding methods.  The sequentially trained vectors performed better across the board, however, which suggests the usefulness of ``priming" in this type of vector learning algorithm.

%It should be noted that while the jointly-trained vectors led to negligible gain in this case, the sequentially trained vectors performed better across the board, with the difference for the MLP classifier edge on the edge of statistical significance\footnote{{\em p-value} = 0.052.} and a much more substantial improvement for the other methods.  This amounts to training on the structural features where the learning algorithm is ``primed" with standard Word2Vec embeddings.  We hypothesize that this is because while the structural features are demonstrably more important than bag-of-words context here, it can only help to start from an embedding space which already has some meaningful characteristics for the task at hand.  

\subsection{Paraphrase Database Lexical Pairs} 

As another source of term-matching data, we used pairs from the small (highest-confidence) lexical paraphrase dataset of the Paraphrase Database \cite{Ganit:13}.  They consist of instances where two distinct words were used to mean the same thing in bi-texts.  In contrast to the WordNet pairs, this dataset skews toward less common words.  It includes many instances of things like alternate spellings and alternate names (or transliterations) for proper nouns.\footnote{The PPDB data consists of 79,696 training pairs, 9,962 validation pairs, and 9,962 test pairs, split 20\%/80\% between matches and non-matches.}

Here, the performance difference between the raw-text and structural-feature vectors was much smaller.  The combined-objective methods were better than either.  This suggests that when the two types of training data have comparable degrees of utility, but work through different avenues, a combined training method can create embeddings which capture the advantages of both approaches.

\section{Conclusion and Future Work}

%It is clear that for some tasks, where the functional role of words is more important than their domain of usage, vectors trained with structural features have advantages over the common bag-of-words approach.  There are likely instances when the reverse is the case as well.  When both properties are desirable, however, as with the described test on paraphrase data, the advantages of each can be captured by using combined-objective training.
%
%The preferred method of doing this also depends on the intended usage.  There is evidence that if both types of features are able to contribute in somewhat equal degree, then parallel, joint training can capture the combined properties of both.  On the other hand, if there is evidence that structural features provide much more direct support for the intended use, gains can still be made by initializing the vector learner with raw-text embeddings.  This could likely be applied to other instances of learning continuous dense representations, were one would sequentially refine vectors, proceeding from more general to more specific features.

It is clear that the best method of training continuous word embeddings depends on the domain of application.  When comparably useful information is encoded in both sentence context and structured representations, combining both as training data can lead to superior vector representations.

For future work, we plan to try using richer feature sets, such as knowledge-base relations, as training inputs, and explore the mathematics behind the sequential training mechanism.

\end{document}